\documentclass[english, usemulticol]{cfasta}
\usepackage{booktabs}
\usepackage{multirow}
\renewcommand{\thefootnote}{\fnsymbol{footnote}}
\usepackage{algorithm}
\usepackage{algorithmic}
\usepackage{threeparttable}
\usepackage{makecell}
\usepackage[comma,numbers,square,sort&compress]{natbib}
\usepackage{epstopdf}
\usepackage{amsmath} % 必须添加，用于 split 环境
\begin{document}

\title{Learning domain-invariant features through channel-level sparsification for Out-Of Distribution Generalization}

\author{Haoran Pei\aref{1},
        Yuguang Yang\aref{3},
        Kexin Liu\aref{1},
        Juan Zhang\aref{2},
        Baochang Zhang\aref{2}}

% 在作者定义结束后立即添加这几行
% \let\thefootnote\relax\footnotetext{$^*$ These authors contributed equally to this work.}
% \let\thefootnote\relax\footnotetext{$^{\dagger}$ Corresponding author.}

% 恢复正常的脚注计数（如果后续还有其他脚注）
\setcounter{footnote}{0}

\affiliation[1]{School of Automation Science and Electrical Engineering, Beihang University, Beijing 100191, P. R. China\email{sy2303515@buaa.edu.cn},\email{kxliu@buaa.edu.cn}}

\affiliation[2]{School of Artificial Intelligence, Beihang University, Beijing 100191, P. R. China\email{bczhang@buaa.edu.cn},\email{zhang\_juan@buaa.edu.cn}}

\affiliation[3]{School of Electronic Information Engineering, Beihang University, Beijing 100191, P. R. China\email{guangbuaa@buaa.edu.cn}}

\maketitle

\begin{abstract}
Out-of-Distribution (OOD) generalization has become a primary metric for evaluating image analysis systems. Since deep learning models tend to capture domain-specific context, they often develop shortcut dependencies on these non-causal features, leading to inconsistent performance across different data sources. Current techniques, such as invariance learning, attempt to mitigate this. However, they struggle to isolate highly mixed features within deep latent spaces. This limitation prevents them from fully resolving the shortcut learning problem.In this paper, we propose Hierarchical Causal Dropout (HCD), a method that uses channel-level causal masks to enforce feature sparsity. This approach allows the model to separate causal features from spurious ones, effectively performing a causal intervention at the representation level. The training is guided by a Matrix-based Mutual Information (MMI) objective to minimize the mutual information between latent features and domain labels, while simultaneously maximizing the information shared with class labels.To ensure stability, we incorporate a StyleMix-driven VICReg module, which prevents the masks from accidentally filtering out essential causal data. Experimental results on OOD benchmarks show that HCD performs better than existing top-tier methods.
\end{abstract}


% Please remove or comment out the following line if the footnote is not necessary

% \footnotetext{This work is supported by National Natural Science
% Foundation (NNSF) of China under Grant 00000000.}

\section{Introduction}

Deep learning models have achieved remarkable success in image analysis. However, their performance often degrades significantly when deployed in unseen domains where the data distribution differs from the training set—a challenge known as Out-of-Distribution (OOD) generalization~\cite{liu2021towards,yang2024limits}. In real-world applications, these distribution shifts typically arise from variations in environmental factors, such as lighting, background context, or sensor specifications~\cite{amerikanos2022image}. For a model to be reliable, it must remain robust to these variations while focusing on the underlying invariant semantic features.

To improve generalization, existing methods generally focus on either data-level or feature-level adjustments. Data-level techniques aim to obscure domain-specific traits by simulating environmental diversity through augmentation. Feature-level frameworks instead focus on maintaining statistical consistency across different training environments. However, both approaches often treat latent features as entangled representations, failing to explicitly separate intrinsic causal factors from spurious correlations. As a result, models remain prone to shortcut learning, where they rely on unstable environmental cues rather than stable semantic features.

Recently, causality-inspired methods have been introduced to mitigate these biases through spatial interventions~\cite{pernice2023out,mao2021generative}. While these methods help direct model attention, they primarily operate in the pixel space, which is often insufficient for resolving entanglement in high-dimensional semantic spaces. In complex visual data, domain biases—such as capture device signatures or environmental noise—are frequently encoded across feature channels rather than being localized in specific pixels. This limitation necessitates a transition from spatial-level masking to more effective representation-level interventions.

To resolve this bottleneck, we propose \textbf{Hierarchical Causal Dropout (HCD)}, a framework that shifts the intervention from the pixel space to the internal representation space. Unlike previous methods, HCD treats feature channels as the fundamental units of intervention, performing structural surgery on the latent manifold. Specifically, we introduce an Advanced Feature Gater that acts as a causal filter, dynamically identifying and suppressing channels entangled with environmental biases. To guide this gating process, we utilize a Matrix-based Mutual Information (MMI) objective~\cite{yu2018multivariate}. By calculating the Von Neumann entropy of feature kernels, we minimize the dependency between the gated representations and domain labels, effectively weaken the non-causal noise from the latent space.

Furthermore, to ensure that the suppression of biased channels does not compromise the preservation of essential task-relevant signals, we integrate a StyleMix-driven VICReg mechanism~\cite{bardes2021vicreg}. By generating synthetic OOD features via style-mixing in the latent space and applying Variance-Invariance-Covariance regularization, HCD forces the model to maintain representation consistency across diverse simulated environments. This structured approach effectively decouples stable causal mechanisms from unstable noise across the network hierarchy.

Finally, we evaluate HCD on diverse OOD benchmarks, including Camelyon17 and iWildCam, covering medical imaging, wildlife monitoring, and general object recognition. Experimental results show that HCD consistently outperforms state-of-the-art methods.

Our primary contributions are summarized as follows:
\begin{itemize}
\item \textbf{Representation-Level Intervention:} We propose Hierarchical Causal Dropout (HCD), a method that uses a learnable gating mechanism for channel-wise causal masking. This approach moves beyond pixel-level perturbations to directly intervene on the latent manifold.
\item \textbf{Information-Theoretic Decoupling:} We introduce a Matrix-based Mutual Information (MMI) objective. By leveraging matrix entropy, this objective quantifies and minimizes the leakage of domain-specific information, effectively isolating stable causal features.
\item \textbf{Style-Invariant Regularization:} We integrate StyleMix-driven VICReg to enforce feature-level consistency. This mechanism ensures the model maintains representation invariance against synthetic distribution shifts, focusing on task-relevant semantic features rather than environmental noise.
\end{itemize}

\section{Related Work}

\textbf{Targeted Feature Masking and Intervention.} 
The evolution of masking techniques has moved from stochastic to structured approaches. Conventional Dropout aims to prevent co-adaptation by randomly zeroing neurons, but it is often inefficient for OOD tasks because it ignores the semantic role of different feature dimensions \cite{wu2021r, mou2018dropout}. To address this, structured masking methods—such as Spatial Dropout or attention-guided masking—were developed to suppress redundant spatial regions rather than individual pixels \cite{guo2023place, li2023rethinking}. Unlike previous spatial-based interventions, HCD identifies and sparsifies specific feature channels that act as carriers for domain-specific noise. This provides a more granular decoupling of causal representations from environmental biases.

\textbf{Information-Theoretic Decoupling and Bias Suppression.} 
A key strategy in mitigating spurious correlations is to minimize the mutual information between latent representations and domain identifiers. While adversarial training is commonly used for this purpose, it often suffers from training instability and sensitive hyperparameter tuning. Recent advances in Information Bottleneck (IB) theory and Mutual Information (MI) estimation \cite{qi2020visualizing, zhang2022towards} provide a more principled framework for feature decoupling. Unlike traditional estimation, HCD leverages Matrix-based Mutual Information (MMI) to penalize domain entanglement, directly constraining the latent manifold at minimal cost to suppress domain-specific information within the learned embeddings.

\textbf{Invariance Learning via Latent Augmentation and Regularization.}
Data-level augmentation often fails to capture the complex distribution shifts present in real-world scenarios. To address this, latent space perturbation methods—such as MixStyle \cite{zhou2021domain} and DSU \cite{li2022uncertainty}—have been developed to simulate domain variations by manipulating feature statistics. MixStyle generates synthetic domains through the interpolation of feature moments, while DSU models these statistics as probabilistic distributions to capture domain uncertainty. However, simply simulating shifts is often insufficient if the resulting latent representations remain redundant or entangled \cite{zhang2022towards, abbasi2024deciphering}. Without explicit constraints, models may still capture spurious correlations through correlated feature pathways despite the increased diversity in the training distribution. Unlike these approaches that focus primarily on simulating distribution shifts, HCD incorporates VICReg \cite{bardes2021vicreg} as an active covariance-based regularization to penalize cross-channel redundancy at minimal cost, thereby ensuring the model captures only invariant semantic features.

\begin{figure*}[t]  % [t] 表示强制放置在页顶 (top)
  \centering
  \includegraphics[width=0.85\textwidth]{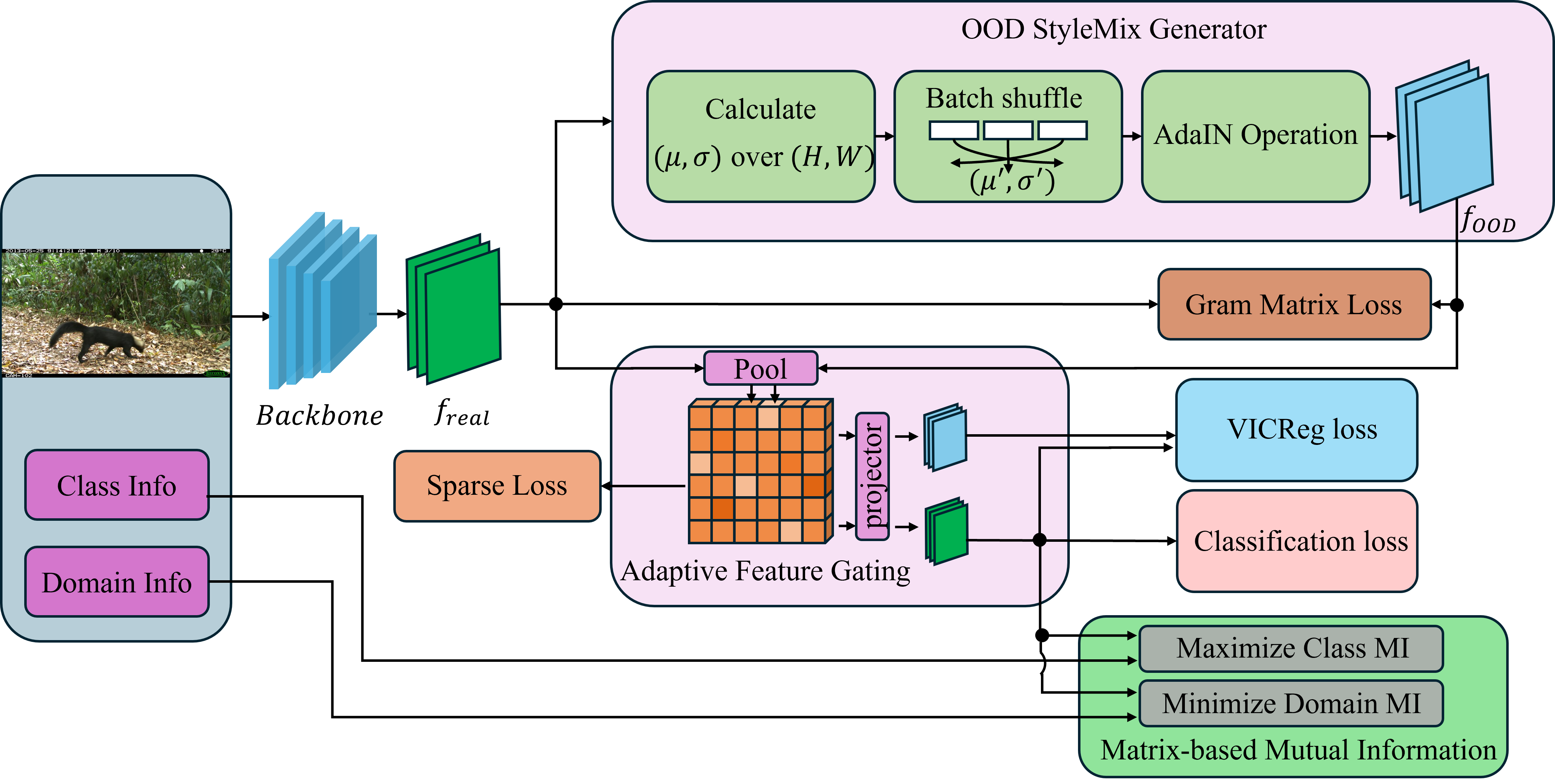} 
  \caption{Overview of the proposed HCD framework. Black dots ($\bullet$) denote branching points.}
  \label{fig:framework}
\end{figure*}

\section{Methodology}
We propose the Hierarchical Causal Dropout (HCD) framework to achieve representation-level decoupling for robust image analysis. In real-world settings, deep learning models often fail under distribution shifts by relying on spurious, non-causal correlations—such as capture noise or environmental variations—rather than stable semantic features.

To resolve feature entanglement, HCD performs representation-level interventions on internal feature channels. As illustrated in Fig. \ref{fig:framework}, the framework consists of three components: (1) Channel-Level Sparsification, which introduces an information bottleneck to isolate domain-specific shortcuts; (2) Information-Theoretic Decoupling, which utilizes matrix-based Rényi information \cite{acharya2016estimating} to minimize domain correlations within the latent manifold; and (3) Latent-Level Regularization, which leverages MixStyle and VICReg to improve OOD generalization.

\subsection{Channel-Level Sparsification for Causal Disentanglement}

In standard convolutional networks, the high-dimensional latent representation $\mathbf{z} \in \mathbf{R}^D$ typically acts as a dense, entangled descriptor. Causal semantic factors (domain-invariant) and spurious environmental factors (domain-dependent) are highly entangled across channel dimensions. Because domain-specific styles are often easier for the network to optimize, the model naturally forms shortcut pathways, where these non-causal factors dominate the final predictions. 

To address this, we introduce an Adaptive Feature Gating Module $\mathcal{G}(\cdot)$ designed to perform representation-level intervention via channel sparsification. Given an intermediate feature map, we apply global average pooling to obtain the latent vector $\mathbf{z}$. The gater subsequently generates a continuous, channel-wise intervention mask $\tilde{\mathbf{m}} \in (0, 1)^D$:

\begin{equation}
\label{eq1}
\tilde{\mathbf{m}} = \sigma\left(\mathbf{W}_2 \cdot \text{ReLU}(\text{BN}(\mathbf{W}_1 \mathbf{z}))\right)
\end{equation}

where $\mathbf{W}_1 \in \mathbf{R}^{\frac{D}{r} \times D}$ and $\mathbf{W}_2 \in \mathbf{R}^{D \times \frac{D}{r}}$ are learnable parameters. The reduction ratio $r$ forces the feature projection through an information bottleneck. This constraint induces a competition mechanism among channels, compelling the network to prune redundant dimensions and assign high activation values exclusively to the most informative pathways. The sparse representation is then formulated as $\mathbf{\hat{z}} = \mathbf{z} \odot \tilde{\mathbf{m}}$, where $\odot$ denotes the Hadamard product.

Sparsification serves as the structural prerequisite for causal disentanglement. By restricting the available channel capacity, the gater forces the representation into a constrained state. When combined with the subsequent information-theoretic penalties (detailed in Section \ref{sec:Info}), the model is encouraged to allocate its limited active channels exclusively to robust, invariant semantic features, effectively excluding domain-specific noise that can no longer fit within the sparse bandwidth.

Furthermore, to prevent the model from developing a excessive reliance on a single dominant channel, we apply a probabilistic dropout layer ($p=0.2$) following the gating mechanism. By injecting a Bernoulli noise vector $\boldsymbol{\xi} \sim \text{Bern}(1-p)$, the expected classification risk becomes:

\begin{equation}
\label{eq2}
\min_\theta \mathbf{E}_{\mathbf{z} \sim P(\mathbf{z}), \boldsymbol{\xi} \sim \text{Bern}(1-p)} [\mathcal{L}_{CE}(f(\mathbf{z} \odot \tilde{\mathbf{m}} \odot \boldsymbol{\xi}), y)]
\end{equation}

This randomized sparsification penalizes single-channel reliance, forcing the predictor $f(\cdot)$ to discover and maintain multiple independent causal pathways. Consequently, even if primary channels are suppressed during training, the remaining active pathways preserve semantic integrity, enhancing the model's OOD robustness.

\subsection{Information-Theoretic Decoupling via Matrix Mutual Information}\label{sec:Info}

To ensure that the gated representation $\mathbf{\hat{z}}$ is independent of the domain identifier $d$ while remaining predictive of the label $y$, we introduce an objective based on Matrix-based Mutual Information (MMI). The theoretical foundation for using MI as a decoupling constraint lies in its ability to measure the statistical dependence between variables. Formally, $I(\mathbf{\hat{z}}; d) = 0$ if and only if $P(\mathbf{\hat{z}}, d) = P(\mathbf{\hat{z}})P(d)$, implying that the distribution of learned features is identical across different domains. 

By minimizing $I(\mathbf{\hat{z}}; d)$, we impose a variational constraint on the latent manifold that penalizes any encoding of domain-specific signatures. In our implementation, this is achieved by minimizing the spectral overlap between the feature kernel matrix $\mathbf{K}_{\hat{z}}$ and the domain kernel matrix $\mathbf{K}_d$. From an information-theoretic perspective, this process acts as a selective filter: it forces the network to discard information that is highly correlated with the environment but redundant for the classification task. 

To implement this without explicit density estimation, we utilize the spectral properties of kernel matrices in a Reproducing Kernel Hilbert Space (RKHS). We define a normalized kernel matrix $\mathbf{K} \in \mathbf{R}^{n \times n}$, where $K_{ij} = \frac{1}{n} \kappa(z_i, z_j)$, which is normalized by its trace to ensure the eigenvalues sum to unity. Following the matrix-based formulation of Rényi entropy \cite{acharya2016estimating} of order $\alpha=2$, the entropy is computed as $S(\mathbf{K}) = -\log_2(\text{tr}(\mathbf{K}^2))$. The objective to minimize Domain MI is then formulated as the loss term $\mathcal{L}_{MI\_D}$:
\begin{equation}
\label{eq3}
\begin{split}
    \mathcal{L}_{MI\_D} &= I(\mathbf{\hat{z}}; d) \\
    &= S(\mathbf{K}_{\hat{z}}) + S(\mathbf{K}_d) - S\left(\frac{\mathbf{K}_{\hat{z}} \odot \mathbf{K}_d}{\text{tr}(\mathbf{K}_{\hat{z}} \odot \mathbf{K}_d)}\right)
\end{split}
\end{equation}
While minimizing $I(\mathbf{\hat{z}}; d)$ bleaches non-causal noise, we simultaneously maximize Class MI to ensure that task-relevant semantic information is preserved, which is defined as $\mathcal{L}_{MI\_C} = -I(\mathbf{\hat{z}}; y)$. 

Furthermore, to encourage the Adaptive Feature Gating module to select only the most discriminative channels and discard redundant features, we impose a Sparse Loss on the gating mask $\mathbf{m}$. This is formulated as the $L_1$-norm of the mask activations:
\begin{equation}
\label{eq4}
    \mathcal{L}_{sparse} = \frac{1}{C} \|\mathbf{m}\|_1
\end{equation}
This joint optimization of mutual information and feature sparsity creates an information bottleneck that converges toward an invariant and discriminative representation.

\subsection{StyleMix-driven VICReg Regularization}

While the Matrix MI-based decoupling acts as a corrective filter against observed domain biases, it risks over-suppressing subtle causal signals that are statistically correlated with the source environment. To mitigate this, we integrate StyleMix latent perturbation with Variance-Invariance-Covariance Regularization (VICReg) to serve as a regularization anchor. This combination enables StyleMix to proactively expand the diversity of the latent manifold through synthetic style perturbations, while VICReg provides the structural stabilization needed to anchor the gated representations against collapse and preserve invariant semantic integrity.

Following the principles of Adaptive Instance Normalization (AdaIN) \cite{huang2017arbitrary}, a feature map $F \in \mathbf{R}^{C \times H \times W}$ can be decomposed into domain-invariant content $\hat{F}$ and domain-specific style statistics $\{\mu(F), \sigma(F)\}$, such that $F = \sigma(F) \cdot \hat{F} + \mu(F)$. To simulate potential domain shifts without accessing target data, our StyleMix Generator implements an AdaIN-based virtual sampling mechanism in the latent space. For a given sample $F_i$, the module shuffles the style statistics within a mini-batch using a random permutation $\pi$, generating a novel perturbed feature $\tilde{F}_i$:
\begin{equation}
\label{eq5}
\begin{split}
\tilde{F}_i = \sigma_{\pi(i)} \cdot \left( \frac{F_i - \mu_i}{\sigma_i + \epsilon} \right) + \mu_{\pi(i)}
\end{split}
\end{equation}
where $\epsilon$ is a small constant used for numerical stability. This operation effectively leverages the AdaIN operator to explore the combinatorial space of style distributions, expanding the observable domain boundaries.

To enforce representation consistency across these synthetic shifts, the VICReg framework is applied to the projected embeddings $z = g(\hat{z})$ and $\tilde{z} = g(\tilde{z}_{aug})$. As defined in our vicreg loss module, the objective incorporates three distinct constraints: invariance, variance, and covariance. The invariance loss $\mathcal{L}_{sim} = \|z - \tilde{z}\|^2_2$ penalizes variations caused by style perturbation, effectively driving the partial derivatives of the representation with respect to style statistics towards zero. In information-theoretic terms, this ensures that $\lim_{\mathcal{L}_{sim} \to 0} I(z; S | C) = 0$, guaranteeing distributional invariance. Simultaneously, the variance constraint $\mathcal{L}_{std} = \frac{1}{D} \sum_{j=1}^{D} \max(0, \gamma - \sigma_j)$ maintains informational richness by forcing the batch-wise standard deviation $\sigma_j$ of each feature dimension above a fixed threshold $\gamma$, while the covariance loss $\mathcal{L}_{cov} = \frac{1}{D} \sum_{i \neq j} [\mathbf{C}(z)]_{i,j}^2$ penalizes the off-diagonal elements of the covariance matrix $\mathbf{C}(z)$ to reduce cross-channel redundancy. Finally, these components are aggregated to form the total VICReg loss: \begin{equation}
\label{eq6}
\begin{split}
\mathcal{L}_{vic} = \lambda_{sim}\mathcal{L}_{sim} + \lambda_{std}\mathcal{L}_{std} + \lambda_{cov}\mathcal{L}_{cov}
\end{split}
\end{equation}
By minimizing $\mathcal{L}_{Gram} = \| \mathbf{GM}(F) - \mathbf{GM}(\tilde{F}) \|^2_F$, we ensure that intrinsic semantic co-activations remain robust against stylistic permutations. This hierarchical approach ensures that both global embeddings and local spatial patterns contribute to OOD robustness.

Finally, the proposed formulation is justified by the domain adaptation theory \cite{ben2010theory}, which bounds the target domain error $\epsilon_{\mathcal{T}}(h)$ by the source error $\epsilon_{\mathcal{S}}(h)$ and the domain divergence $d_{\mathcal{H}\Delta\mathcal{H}}$. Through StyleMix perturbation, we artificially expand the support of the source distribution $\mathcal{D}_{\mathcal{S}}$. Concurrently, the integrated VICReg and Gram constraints explicitly minimize $d_{\mathcal{H}\Delta\mathcal{H}}$ by aligning causal feature distributions across stylistic shifts. This mathematical framework tightens the upper bound of the target risk, providing a theoretical guarantee for superior OOD generalization.

\subsection{Joint Optimization and Curriculum Scheduling}

Following the optimization procedure outlined in Algorithm~\ref{alg:hcd_principle}, the total objective function for HCD explicitly integrates the six core components, formulated as:
\begin{equation}
\label{eq7}
\begin{split}
    \mathcal{L}_{total} =& \mathcal{L}_{cls} + \beta_1 \mathcal{L}_{vic} + \beta_2 \mathcal{L}_{Gram} \\
    &+ \beta_{3}^{(t)} \mathcal{L}_{MI\_C} + \beta_{4}^{(t)} \mathcal{L}_{MI\_D} + \beta_{5}^{(t)} \mathcal{L}_{sparse}
\end{split}
\end{equation}
To effectively balance these objectives and ensure training stability, we implement the weight coefficients $\beta_{3}^{(t)}$, $\beta_{4}^{(t)}$, and $\beta_{5}^{(t)}$ using a curriculum scheduling strategy. 

Specifically, in the initial epochs, the model focuses on learning basic discriminative features by setting these penalty weights near zero. As training progresses, they are gradually increased to their target values. This curriculum allows the model to transition smoothly from a standard classifier to a domain-agnostic feature extractor, preventing the sparsification and disentanglement mechanisms from collapsing before the network has captured sufficient semantic information.
\begin{algorithm}[H]
\caption{Hierarchical Causal Dropout (HCD) Framework}
\label{alg:hcd_principle}
\begin{algorithmic}[1]
\STATE \textbf{Input:} Training data $X$, category labels $y$, and domain indices $d$
\STATE \textbf{Output:} Optimized domain-invariant model $h(f(\cdot))$
\STATE Initialize model parameters $\theta$ and feature gating module $G(\cdot)$
\FOR{each training iteration}
    \STATE \textbf{Feature Selection \& Encoding:}
    \STATE Extract backbone features: $F = f(X)$
    \STATE Apply channel-wise gating for task-relevant selection: $z, mask = G(F)$
    \STATE \textbf{Stylized Feature Synthesis:}
    \STATE Generate OOD variations via StyleMix (statistic shuffling): $F_{style} = \text{StyleMix}(F)$
    \STATE Encode stylized features: $z_{style} = G(f(F_{style}))$
    \STATE \textbf{Loss Computation (Core Principles):}
    \STATE 1. \textit{Classification}: $\mathcal{L}_{cls} = \text{CrossEntropy}(h(z), y)$
    \STATE 2. \textit{Sparsity}: $\mathcal{L}_{sparse} = \text{L1Norm}(mask)$
    \STATE 3. \textit{Style Regularization}: $\mathcal{L}_{vic}$ and $\mathcal{L}_{Gram}$
    \STATE 4. \textit{Information Bottleneck}: $\mathcal{L}_{MI\_C}$ and $\mathcal{L}_{MI\_D}$
    \STATE \textbf{Total Objective:}
    \STATE $\mathcal{L} = \mathcal{L}_{cls} + \beta_1 \mathcal{L}_{vic} + \beta_2 \mathcal{L}_{Gram} + \beta_3 \mathcal{L}_{MI\_C} + \beta_4 \mathcal{L}_{MI\_D} + \beta_5 \mathcal{L}_{sparse}$
    \STATE Update $\theta$ by minimizing $\mathcal{L}$
\ENDFOR
\STATE \textbf{return} Optimized model $h(f(\cdot))$
\end{algorithmic}
\end{algorithm}

\section{EXPERIMENTAL SETUP}

\subsection{Datasets and Implementation Details}
We evaluate the HCD framework on two large-scale benchmarks from the WILDS collection\cite{koh2021wilds}, which represent distinct and challenging real-world distribution shifts:
\begin{itemize}
    \item Camelyon17\cite{bandi2018detection}: A digital pathology dataset for tumor detection in lymph node sections. It captures clinical distribution shifts across five medical centers. Variations in staining protocols and imaging equipment present a formidable challenge for model robustness.
    \item iWildCam\cite{beery2021iwildcam}: A wildlife monitoring dataset where the model must generalize to unseen camera trap locations. The shift involves drastic variations in illumination, background vegetation, and sensor characteristics across 323 different sites.
\end{itemize}

\textbf{Architectural Configuration.} To ensure a fair comparison, we adopt DenseNet-121 as the backbone for Camelyon17 and ResNet-50 for iWildCam, following standard benchmark protocols. In contrast to many existing works that leverage ImageNet-pretrained weights, our model is trained from scratch to ensure that the learned representations are derived entirely from the target histopathological and ecological domains, avoiding pre-training bias. 

\textbf{Optimization and Hyperparameters.} We utilize the Adam optimizer with a learning rate of $2 \times 10^{-4}$ and a batch size of 32. To maintain training stability and prevent premature collapse of the disentanglement mechanism, we strictly follow the curriculum scheduling strategy. Specifically, the domain-disentanglement weight $\beta_3^{(t)}$ is initially set to $0.5$ and scaled to $5.0$ after the first epoch, while the sparsity penalty $\beta_4^{(t)}$ is gradually increased from $0.005$ to $0.1$ starting from the third epoch ($t \geq 2$). For the representation consistency terms, we fix $\lambda_{v}=1.0$ and $\beta=1.0$. All experiments are conducted across five independent runs with fixed random seeds (0--4) to guarantee reproducibility.

\subsection{Comparative Baselines}
We compare HCD against two representative paradigms in domain generalization and efficient learning:
\begin{itemize}
    \item \textbf{ERM (Empirical Risk Minimization)\cite{vapnik1991principles}:} The standard training baseline that minimizes the average loss across all training domains without explicit domain-invariant regularization.
    \item \textbf{Bonsai \cite{zhang2022rich}:} A state-of-the-art method for resource-efficient learning that employs tree-based conditional computation. We use it as a primary competitor to evaluate how our gating-based sparsification compares to traditional sparse learning approaches in the context of domain decoupling.
\end{itemize}

Using the features derived from these models as a foundation, we extended our evaluation to incorporate several prominent domain-invariant strategies, such as IRMv1\cite{arjovsky2019invariant}, IRMX\cite{chen2022pareto}, VREx\cite{krueger2021out}, and GroupDRO\cite{sagawa2019distributionally}.

\subsection{Representation Analysis and Interpretability Protocols}

To evaluate the decision-making logic and generalization robustness of the HCD framework, we implement two complementary visualization protocols:

\textbf{Spatial Attribution via CAM.} Following a comparative analysis with representative class activation mapping methods, such as CAM \cite{zhou2016learning} and DecomCAM\cite{yang2024decomcam}, we employ Grad-CAM \cite{selvaraju2017grad} for decision visualization, primarily driven by its superior architectural flexibility. While CAM is limited by specific structural constraints and DecomCAM introduces higher computational overhead, Grad-CAM produces high-fidelity saliency maps for arbitrary CNN architectures without necessitating model modifications, thereby ensuring a both faithful and efficient interpretation. By generating heatmaps from the final convolutional layers, we verify whether HCD directs its attention toward invariant semantic structures (e.g., pathological markers or animal silhouettes) rather than domain-specific environmental noise (e.g., staining artifacts or background textures). This provides direct visual evidence that our disentanglement mechanism effectively filters out spurious correlations.

\textbf{Generalization Analysis via Loss Landscape.} We analyze the Loss Landscape Sharpness \cite{foret2020sharpness} to assess the model's stability against distribution shifts, as the loss landscape provides an intuitive way to demonstrate the model's robustness. By visualizing the loss surface around the converged parameters, we aim to demonstrate that HCD leads the model into a flatter optimization region compared to ERM. This relative flatness reflects the model's stability under the interference of domain-specific information; in optimization theory, such stability is strongly correlated with superior OOD performance and reduced sensitivity to the heterogeneous context shifts encountered in unseen domains, ensuring more reliable performance across diverse environments.

\section{RESULTS AND ANALYSIS}
\begin{table}[!htb]
  \centering
  \caption{Camelyon and iWildCam Results}
  \label{tab:camelyon17_results}
  \begin{tabular}{|l|l|c|c|}
    \hhline
    INIT. & METHOD & Camelyon & iWildCam \\
    \hline\hline
    ERM & IRM & 75.68(±7.41)\% & 28.76(±0.45)\% \\
    ERM & IRMX & 73.49(±9.33)\% & 28.82(±0.47)\% \\
    ERM & GroupDRO & 76.09(±6.46)\% & 28.51(±0.58)\% \\
    ERM & VRex & 71.60(±7.88)\% & 28.82(±0.47)\% \\
    \hline
    Bonsai & IRM & 73.59(±6.16)\% & 27.60(±1.57)\% \\
    Bonsai & IRMX & 64.77(±10.1)\% & 27.62(±0.66)\% \\
    Bonsai & GroupDRO & 72.82(±5.37)\% & 27.16(±1.18)\% \\
    Bonsai & VRex & 76.39(±5.32)\% & 25.81(±0.42)\% \\ 
    \hline
    HCD & IRM & 84.31(±2.64)\% & 32.94(±0.63)\% \\
    HCD & IRMX & 85.91(±1.70)\% & 31.10(±0.48)\% \\
    HCD & GroupDRO & 86.16(±2.97)\% & 33.09(±0.62)\% \\
    HCD & VRex$^*$ & 86.62(±2.65)\% & 31.57(±0.79)\% \\
    \hhline
  \end{tabular}
\end{table}

\textbf{Comparison results.} As summarized in Table~\ref{tab:camelyon17_results}, the proposed HCD framework demonstrates superior generalization performance across both histopathological and ecological monitoring domains. On the Camelyon17 dataset, HCD achieves a peak accuracy of 86.62\%, significantly outperforming the standard ERM and Bonsai baseline. More importantly, in the challenging long-tailed iWildCam benchmark, HCD maintains a robust accuracy range of 31.10\%--33.09\%. This result is particularly significant as it resolves the performance collapse typically observed in spatial-domain perturbation methods when dealing with rare species. Unlike aggressive spatial augmentation, HCD's channel-wise sparsification and information bottleneck mechanisms offer a more essential way to filter environmental noise. By effectively stripping away spurious correlations while preserving sparse but vital semantic features of tail categories, HCD ensures that the model does not sacrifice rare-class discriminability for domain invariance.

\begin{figure}[!htb]
  \centering
  % 1. 使用 \hsize 贴合官方写法
  % 2. 确保你的图片已经转换成 .eps 格式（如果环境支持）
  \includegraphics[width=\hsize]{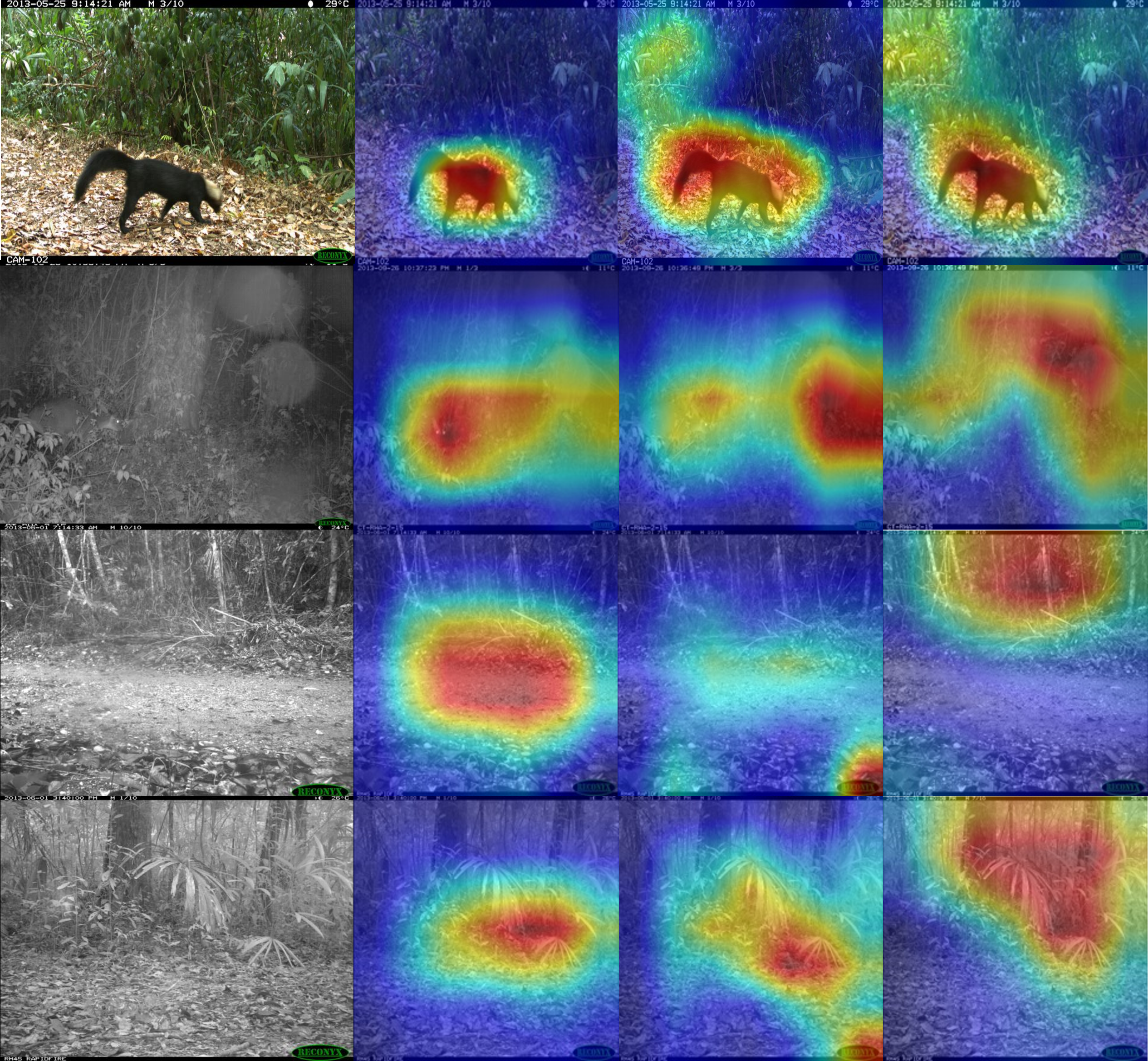} 
  
  % 3. 官方通常要求 9pt 字体，注意不要在标题里加过多粗体
  \caption{Visual comparison of Grad-CAM on the iWildCam dataset. Columns from left to right: HCD, Bonsai, and ERM. HCD maintains precise target localization across complex scenarios such as nocturnal noise, infrared imaging, and severe vegetation occlusion , effectively ignoring domain-specific environmental backgrounds.}
  \label{fig:cam_visualization} % 建议使用简单的标签名
\end{figure}

\textbf{Class Activation Mapping results.} To further interpret the decision-making logic, we analyze the Grad-CAM results on iWildCam, as shown in Fig.~\ref{fig:cam_visualization}. As illustrated in this qualitative analysis, HCD exhibits a remarkable ability to lock onto the invariant semantic core of objects across diverse and degraded conditions, including high-noise nocturnal shots, infrared captures, and heavy vegetation occlusion. While baseline models like Bonsai or ERM often exhibit significant attention dispersion—shifting toward environmental artifacts or background textures—HCD's focus remains tightly anchored to the core contours of the animal subjects. This provides direct visual evidence that our Hierarchical Causal Gating effectively suppresses the activation of shortcut channels that encode domain-specific noise. By physically blocking non-causal information flow, HCD ensures that the network's predictive logic is grounded in stable biological markers rather than spurious environmental correlations.

\begin{figure}[!htb]
  \centering
  % 1. 使用 \hsize 贴合官方写法
  % 2. 确保你的图片已经转换成 .eps 格式（如果环境支持）
  \includegraphics[width=\hsize]{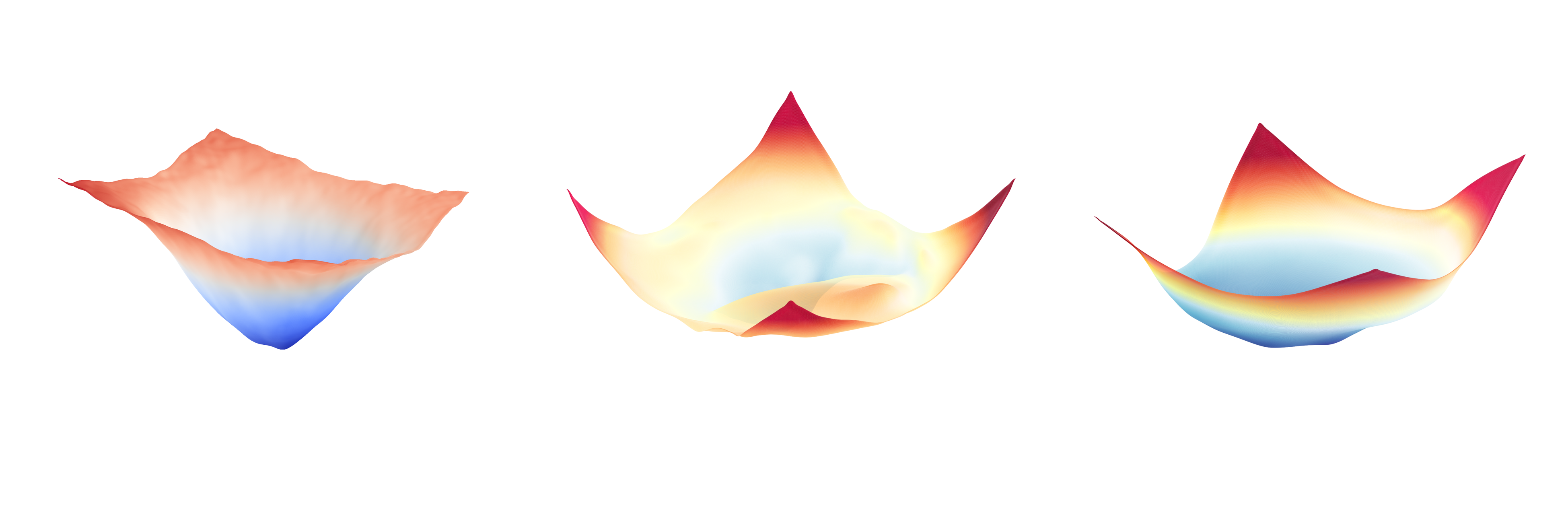} 
  
  % 3. 官方通常要求 9pt 字体，注意不要在标题里加过多粗体
  \caption{Visualization of Loss Landscapes. Columns from left to right: HCD, Bonsai, and ERM. HCD demonstrates a significantly flatter and more expansive basin structure, reflecting its superior stability and reduced sensitivity to the distribution shifts encountered in open-world environments.}
  \label{fig:loss_landscape} % 建议使用简单的标签名
\end{figure}
\textbf{Loss landscapes results.} The robustness of HCD is further corroborated by the Loss Landscape analysis. Our framework yields a notably wider, smoother, and more symmetrical basin structure compared to the sharper landscapes of traditional DG methods. In optimization theory, such flat minima are indicative of a parameter space that is highly insensitive to distribution shifts. By reshaping the topological structure of the parameter space through Matrix MI and sparsity constraints, HCD ensures that both head and tail classes reside in a stable optimization region. This geometric stability explains the framework's consistent performance across unseen hospitals and camera sites, confirming that the model has converged to a solution that prioritizes causal invariance over empirical fitting.

\section{Conclusions}
In this paper, we proposed \textbf{HCD}, a novel representation learning framework that integrates channel sparsification with matrix-based information bottleneck theory to achieve robust Out-of-Distribution generalization. Unlike traditional methods that rely solely on soft regularization, HCD establishes a synergistic mechanism involving Feature Style Mixing for counterfactual intervention, Matrix Rényi Mutual Information for precise disentanglement, and Adaptive Channel Gating for the physical blockage of non-causal information.

Extensive experiments on the Camelyon17 and iWildCam benchmarks demonstrate that HCD significantly outperforms standard ERM and sparse-learning baselines like Bonsai. Notably, HCD exhibits superior stability in long-tailed distributions, as its gentle channel-filtering approach protects scarce semantic features of tail categories while effectively stripping away institutional and environmental biases. Visualizations via CAM and Loss Landscapes further confirm that HCD leads the model toward a flatter optimization region and anchors its attention onto invariant semantic cores across diverse and degraded scenarios.

Despite these gains, certain limitations persist: the current matrix-based MI estimation entails quadratic computational complexity with respect to batch size, which may limit scalability to extremely large-scale training regimes. Future work will focus on developing low-rank approximations for spectral entropy calculation and exploring the applicability of HCD across a broader range of multimodal and self-supervised learning architectures

\balance

\end{document}